\documentclass[letterpaper, 10 pt, conference]{ieeeconf}  

\IEEEoverridecommandlockouts                              

\usepackage{graphicx}
\usepackage{amsmath}
\usepackage{amssymb}
\usepackage{booktabs}
\usepackage{nicefrac}
\usepackage{comment}
\let\labelindent\relax
\usepackage{enumitem}
\usepackage{todonotes}

\usepackage[pagebackref,breaklinks,colorlinks]{hyperref}

\usepackage[capitalize]{cleveref}
\crefname{section}{Sec.}{Secs.}
\Crefname{section}{Section}{Sections}
\Crefname{table}{Table}{Tables}
\crefname{table}{Tab.}{Tabs.}

\title{\LARGE \bf
UAV-Assisted Maritime Search and Rescue: A Holistic Approach
}

\author{Martin Messmer$^{1}$, Benjamin Kiefer$^{1}$, Leon Amadeus Varga$^1$, and Andreas Zell$^{1}$
\thanks{*This work was supported by the German Ministry for Economic Affairs and Climate Action under grant number FKZ: 19A21009C}
\thanks{$^{1}$Faculty of Computer Science, University of Tuebingen, 72076 Tübingen, Germany
        {\tt\small martin.messmer@uni-tuebingen.de}}%
}

\begin{document}

\maketitle
\thispagestyle{empty}
\pagestyle{empty}

\begin{abstract}
    In this paper, we explore the application of Unmanned Aerial Vehicles (UAVs) in maritime search and rescue (mSAR) missions, focusing on medium-sized fixed-wing drones and quadcopters. We address the challenges and limitations inherent in operating some of the different classes of UAVs, particularly in search operations. Our research includes the development of a comprehensive software framework designed to enhance the efficiency and efficacy of SAR operations. This framework combines preliminary detection onboard UAVs with advanced object detection at ground stations, aiming to reduce visual strain and improve decision-making for operators. It will be made publicly available upon publication. We conduct experiments to evaluate various Region of Interest (RoI) proposal methods, especially by imposing simulated limited bandwidth on them, an important consideration when flying remote or offshore operations. This forces the algorithm to prioritize some predictions over others.
\end{abstract}


\section{Introduction}

The vast and unpredictable nature of maritime environments presents a significant challenge for search and rescue (SAR) operations. Traditional methods, often reliant on manned vessels and aircraft, face limitations in speed, accessibility, and risk to human life \cite{costguardmanual}. With the advent of Unmanned Aerial Vehicles (UAVs), new possibilities have emerged, offering enhanced efficiency, safety, and endurance in maritime SAR (mSAR) missions. This paper delves into the application of UAVs in mSAR, specifically medium-sized fixed-wing drones and quadcopters, focusing predominantly on their utility in search operations due to their physical constraints.

In particular, we discuss various classes of drones, including quadcopters of different sizes and fixed-wing drones, along with their respective advantages and disadvantages. We delve into how each type of drone could be effectively utilized at different stages of a mSAR mission, and address the potential challenges and limitations when operating them in such scenarios.

One key aspect of our research is the development of a comprehensive software framework that enables the prediction of preliminary detection onboard the UAVs, followed by a more capable but more compute intensive object detector on the interesting regions of the image on a more powerful ground station \cite{carion2020detr, girshick2015fastrcnn}.

\begin{figure}[ht]
    \centering
    \includegraphics[width=0.99\linewidth]{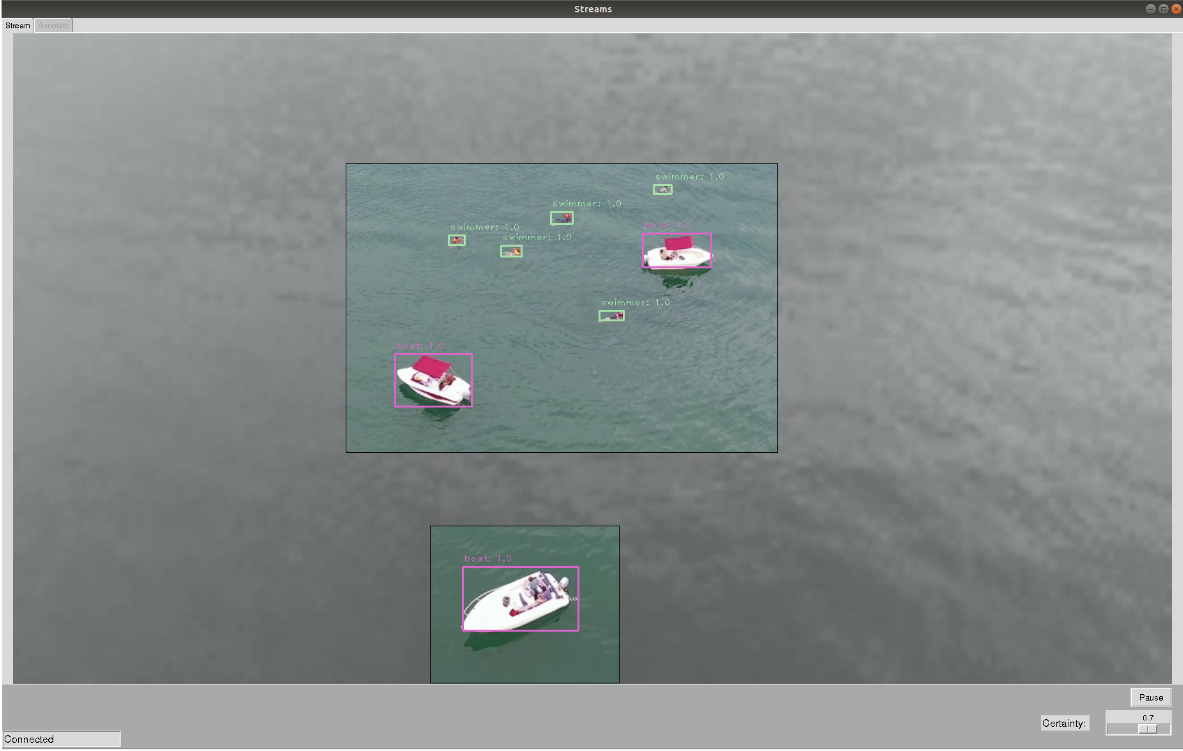}
    \caption{Example footage from the GUI of the ground station software.}
    \label{fig:software_screenshot}
\end{figure}

The initial detections, called Regions of Interest (RoI) \cite{kiefer2023fast}, are intended as the primary focus for both the detection system and the operator at the ground station as they are the most likely to contain humans or vessels in distress.
Additionally, the software is designed to stream these regions of interest in their full quality, in contrast to the rest of the image, which is transmitted in lower quality and gray-scale (see figure \ref{fig:software_screenshot}). This approach accounts for potential constraints of limited bandwidth, which can be a significant factor when operating UAVs in remote locations or far off-shore. 
Therefore, the proposals for regions of interest serve multiple purposes and are a vital component of the overall pipeline. Hence we present an evaluation of various methods for generating these proposals in Section \ref{sec:experiments} of this paper. In particular, we put emphasis on the bandwidth constraint in these experiments, evaluating the performance of each algorithm under various transmission restrictions.

\begin{figure}[t]
    \centering
    \includegraphics[width=0.99\linewidth]{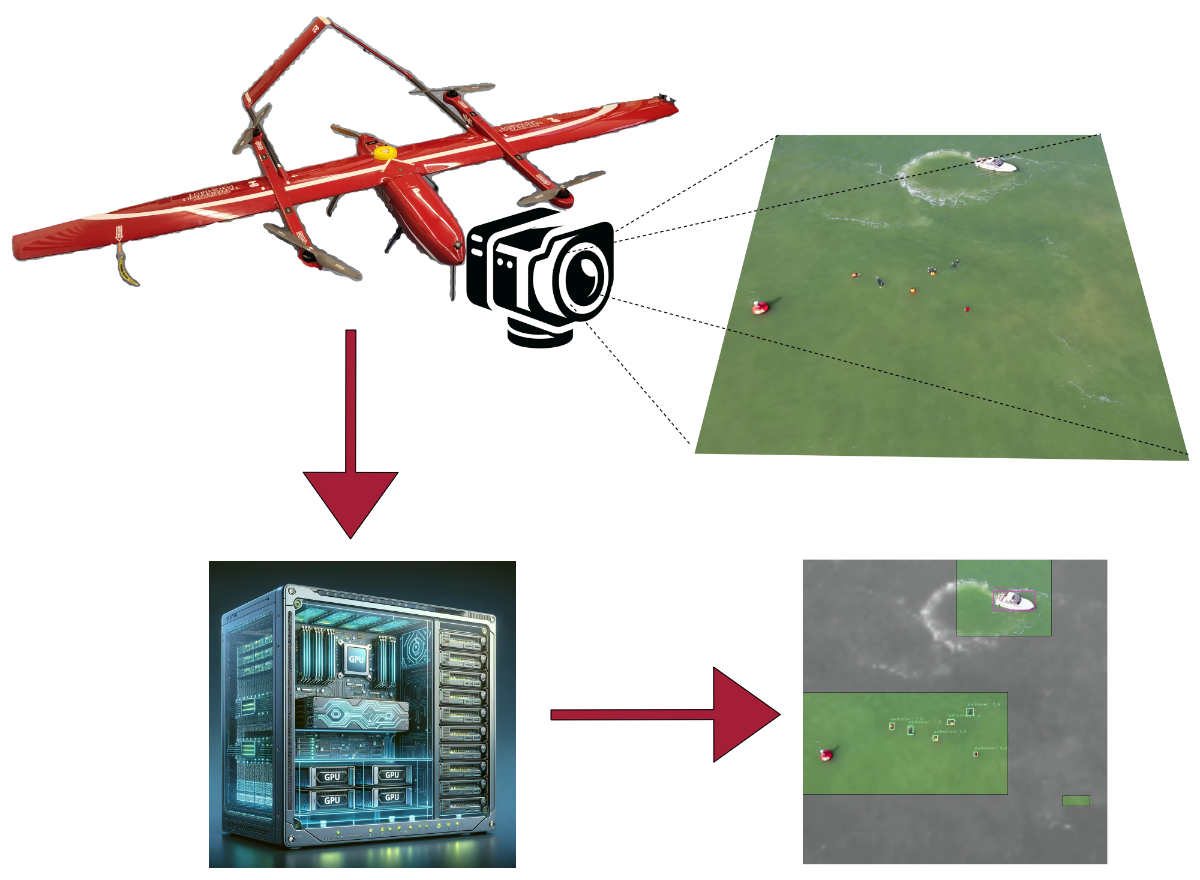}
    \caption{The maritime surface is captured by the drone and then, together with the preliminary detections (RoIs) transmitted to a GPU server on the ground. This then performs fine-grained detection on the RoIs and displays the whole scene to the SAR operator.}
    \label{fig:overview_image}
\end{figure}

By advancing the integration of UAVs in mSAR missions, our research aims to significantly enhance the efficiency and effectiveness of maritime search operations, ultimately contributing to faster response times and increased chances of successful rescue in maritime emergencies. 
Moreover, this software design aims to reduce the visual strain on the drone operator on the ground by automating a portion of the decision-making process. This aids decision making, as watching a screen extensively might lead to 'eye fatigue' \cite{ColesBrennan2019ManagementOD, council2016eyes, sullivan2008visual}, which might potentially lead to bad decisions. Our hope is that this automation will long term lead to operators making more accurate decisions by minimizing the chances of overlooking important details.

In \cite{lygouras2019unsupervised} the authors propose a full system for UAV-based mSAR missions. In contrast to the paper at hand they perform regular object detection fully onboard the drone on a NVIDIA Jetson TX1.
The work \cite{stabernack2021architecture_fpga} concerns itself with the stream of a low-quality video with embedded high-quality regions of interests on FPGAs.
In \cite{mayer2019drones} the authors discuss the advantages of using drones in SAR missions in general which lays the foundations for works like ours.
An interesting RoI proposal method is given in \cite{kiefer2023fast}. However, it requires working with video streams which would limit the choice of data set for testing in this work.

The remainder of the paper is structured as follows: Section \ref{sec:drones} provides an overview of different drone classes, discussing their advantages and disadvantages. Section \ref{sec:software} discusses the software framework we propose, including both, UAV and ground station components. Finally, section \ref{sec:experiments} details our experimental setup and the results and observations obtained. Finally, Section \ref{sec:conclusion} concludes with a discussion of our findings and their implications.

\section{Choosing the Right Drone for mSAR Missions}
\label{sec:drones}

\begin{table*}[t]
\centering
    \begin{tabular}{c|ccc}
                                                                       & \begin{tabular}[c]{@{}c@{}}DJI\\ Matrice 210\end{tabular} & \begin{tabular}[c]{@{}c@{}}Quantum Systems\\ Trinity F90+\end{tabular} & \begin{tabular}[c]{@{}c@{}}ElevonX\\ Skyeye Sierra VTOL\end{tabular} \\\hline\hline
    System                                                             & Quadcopter      & VTOL Fixed Wing                                                        & VTOL Fixed Wing                                                      \\\hline
    Weight                                                             & $5~kg$          & $5~kg$                                                                 & $12.5~kg$                                                            \\\hline
    \begin{tabular}[c]{@{}c@{}}Max. Payload\\ Weight\end{tabular}      & $1.3~kg$        & \checkmark                                                                     & $3~kg$                                                               \\\hline
    \begin{tabular}[c]{@{}c@{}}Air Speed\\ (Range)\end{tabular}        & $0-12~ \nicefrac{m}{s}$        & $17-21~ \nicefrac{m}{s}$                                                              & $17-21~ \nicefrac{m}{s}$                                                            \\\hline
    \begin{tabular}[c]{@{}c@{}}Min. Flight\\ Altitude\end{tabular}     & $0~m$           & $30~m$                                                                 & $50~m$                                                               \\\hline
    \begin{tabular}[c]{@{}c@{}}Max. Flight\\ Time\end{tabular}     & $30~\textit{min.}$           & $90~\textit{min.}$                                                                 & \begin{tabular}[c]{@{}c@{}}$3~h$ (electric) \\ $5~h$ (gas) \end{tabular}                                                          \\\hline
    \begin{tabular}[c]{@{}c@{}}Command \& Control\\ Range\end{tabular} &     $8 ~ km$            &              $7.5 ~ km$                                                          &                     $20 ~ km$                                                 \\\hline
    Wingspan                                              &       $89 ~ cm$          &         $239 ~ cm$                                                               &                                   $300 ~ cm$   \\\hline
    Number of Operators                                              &       $1$          &         $1$                                                               &                                   $2$                                  
    \end{tabular}
    \caption{Technical data of suitable UAVs \cite{m210_datasheet, trinity_datasheet, elevonx_datasheet}.}
    \label{tab:drone_tab}
\end{table*}

\begin{figure}[ht]
    \centering
    \includegraphics[width=0.99\linewidth]{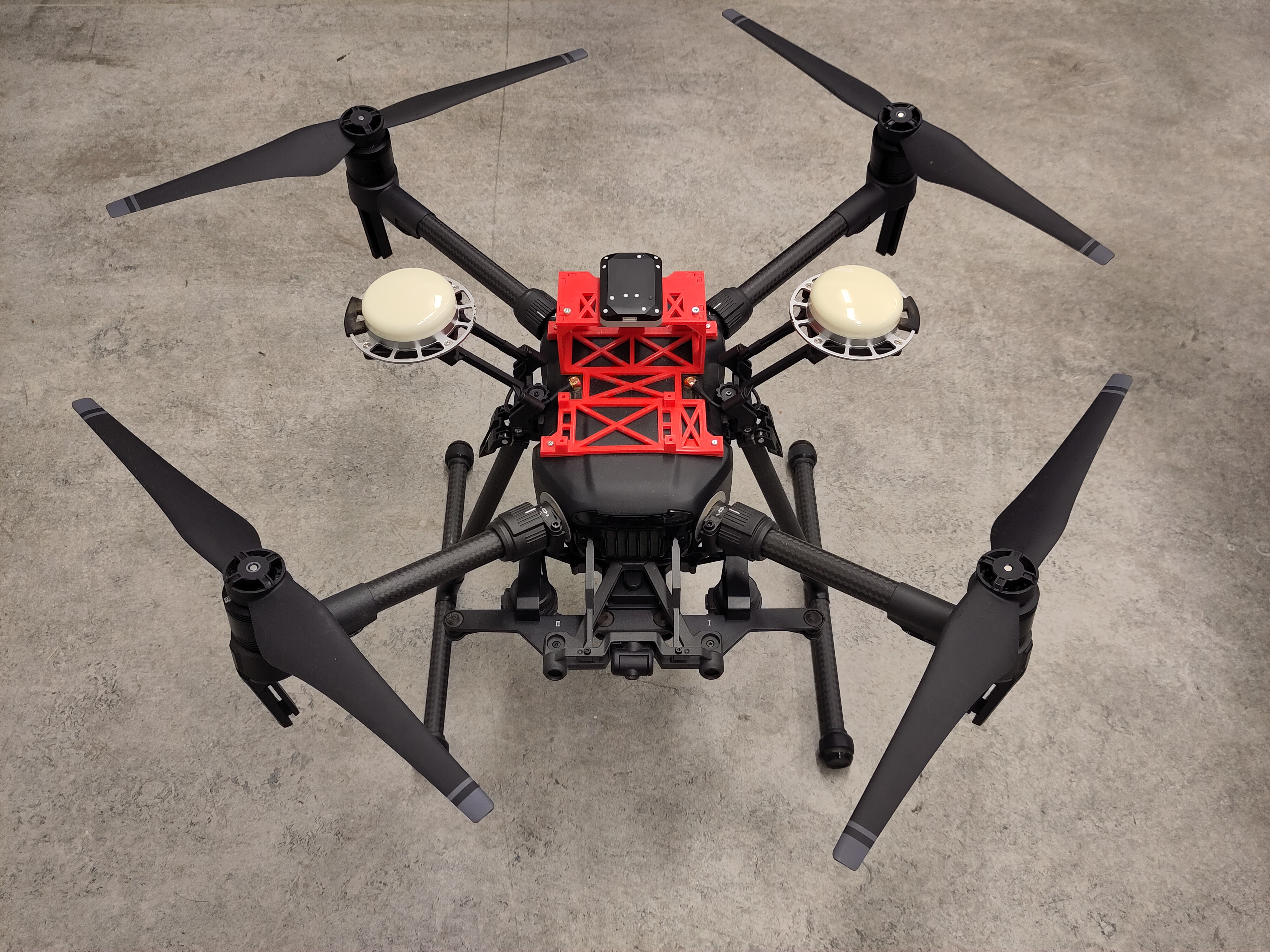}
    \caption{DJI Matrice 210}
    \label{fig:m210}
\end{figure}

Table \ref{tab:drone_tab} shows the technical details of the DJI Matrice M210, the Quantum Systems Trinity F90+, and the ElevonX SkyEye Sierra VTOL. All of these were used in various capacities in our experiments regarding SAR research. Their advantages and disadvantages are meant to be exemplarily for their respective classes of drones. \\
The DJI Matrice 210 (figure \ref{fig:m210}), as a quadcopter, offers excellent maneuverability and the ability to hover in place, which is crucial {for precise, targeted searches and for capturing} specific areas of interest. However, its flight time and speed are less than the other two, which may limit its range and efficiency in covering large maritime areas.

\begin{figure}[ht]
    \centering
    \includegraphics[width=0.99\linewidth]{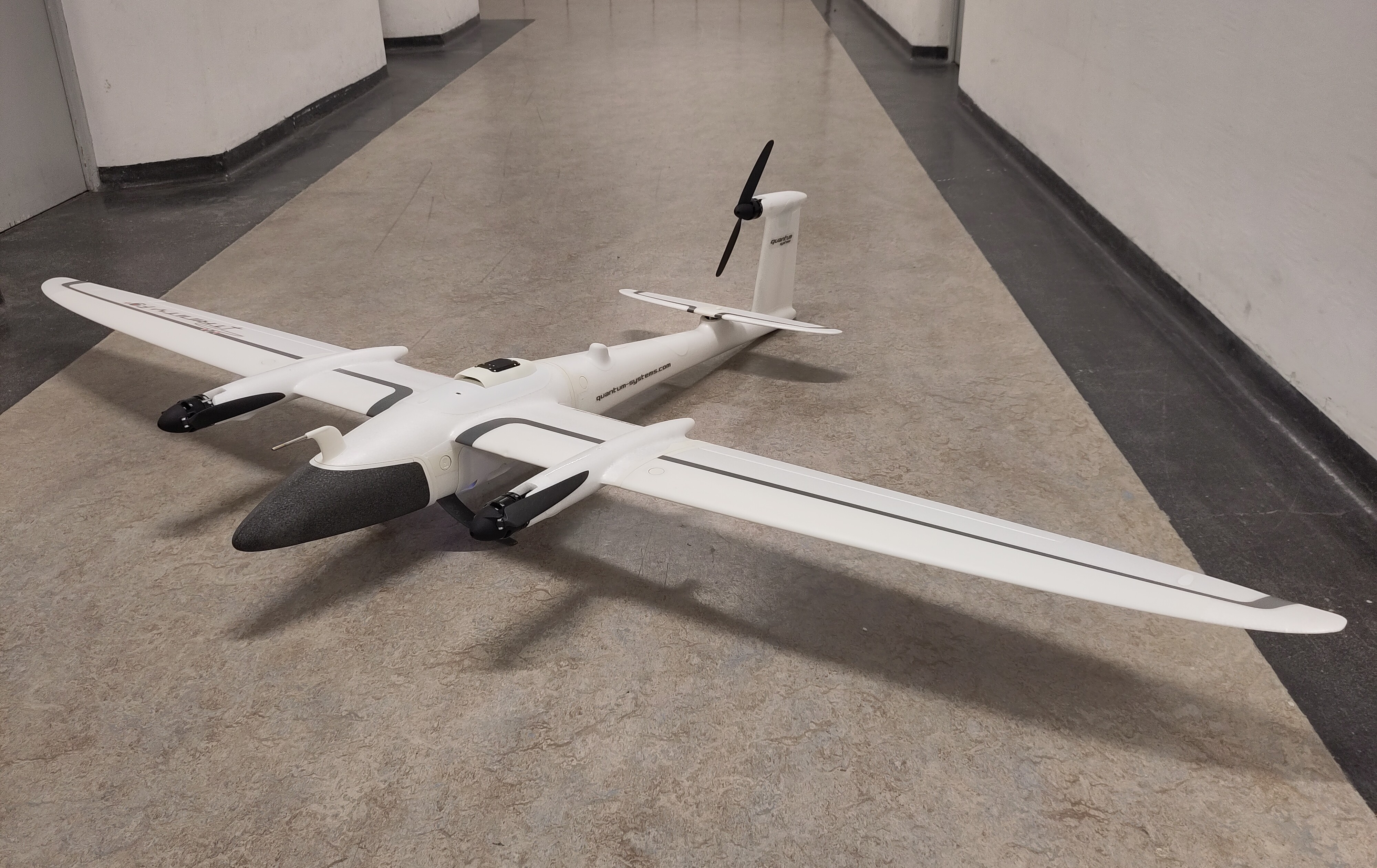}
    \caption{Quantum Systems Trininty F90+}
    \label{fig:trinity}
\end{figure}

The Quantum Systems Trinity F90+ (figure \ref{fig:trinity}) is a fixed-wing VTOL (vertical take-off and landing) that provides a longer flight time of roughly 90 minutes and a large coverage area, making it suitable for initial wide-area searches. Its ability to carry various sensors is advantageous for generating a diverse dataset necessary for training and refining deep learning algorithms. The fixed-wing design offers higher speed and greater efficiency over long distances compared to quadcopters, but it lacks the ability to hover, which can be a limitation for close inspections. Additionally, it significantly complicates the operation of the UAV compared to quadcopters, placing a considerable demand on the operator to maintain a flawless overview of the situation at all times and to think ahead. Furthermore, the maneuverability is limited compared to quadcopters, resulting in an inability to be operated in confined spaces.

One of the key characteristics of the Trinity F90+ is its limited adjustability, especially regarding payload integration. Users can choose from a range of cameras supported by the manufacturer, including multiple different RGB and Multispectral cameras, a LiDAR scanner, and an Oblique RGB camera\cite{trinity_cameras}. The latter has five lenses, each oriented slightly differently, designed for 3D mapping. This selection caters to various data collection needs, from detailed imagery to topographical mapping. However, the inability to integrate custom payloads or sensors outside the manufacturer's offerings restricts the drone's versatility. In particular the inability to employ onboard processing in the form of an NVIDIA Orin \cite{nvidia_orin} or similar.

Furthermore, the Trinity F90+ operates with a proprietary communication system that lacks openness or user adjustability. The absence of a customizable down-link connection means that users are confined to the data transmission and control options provided by the manufacturer. This could pose challenges in integrating the drone into a broader system that employs deep learning algorithms. More precisely, it is impossible to stream captured data instantaneously. This rules out online post-processing on a more powerful GPU server on the ground.

In summary, the Trinity F90+ can play an important role in AI-aided mSAR missions by gathering extensive amounts of high-quality aerial data necessary to train deep learning algorithms used for detection.

\begin{figure}[ht]
    \centering
    \includegraphics[width=0.99\linewidth]{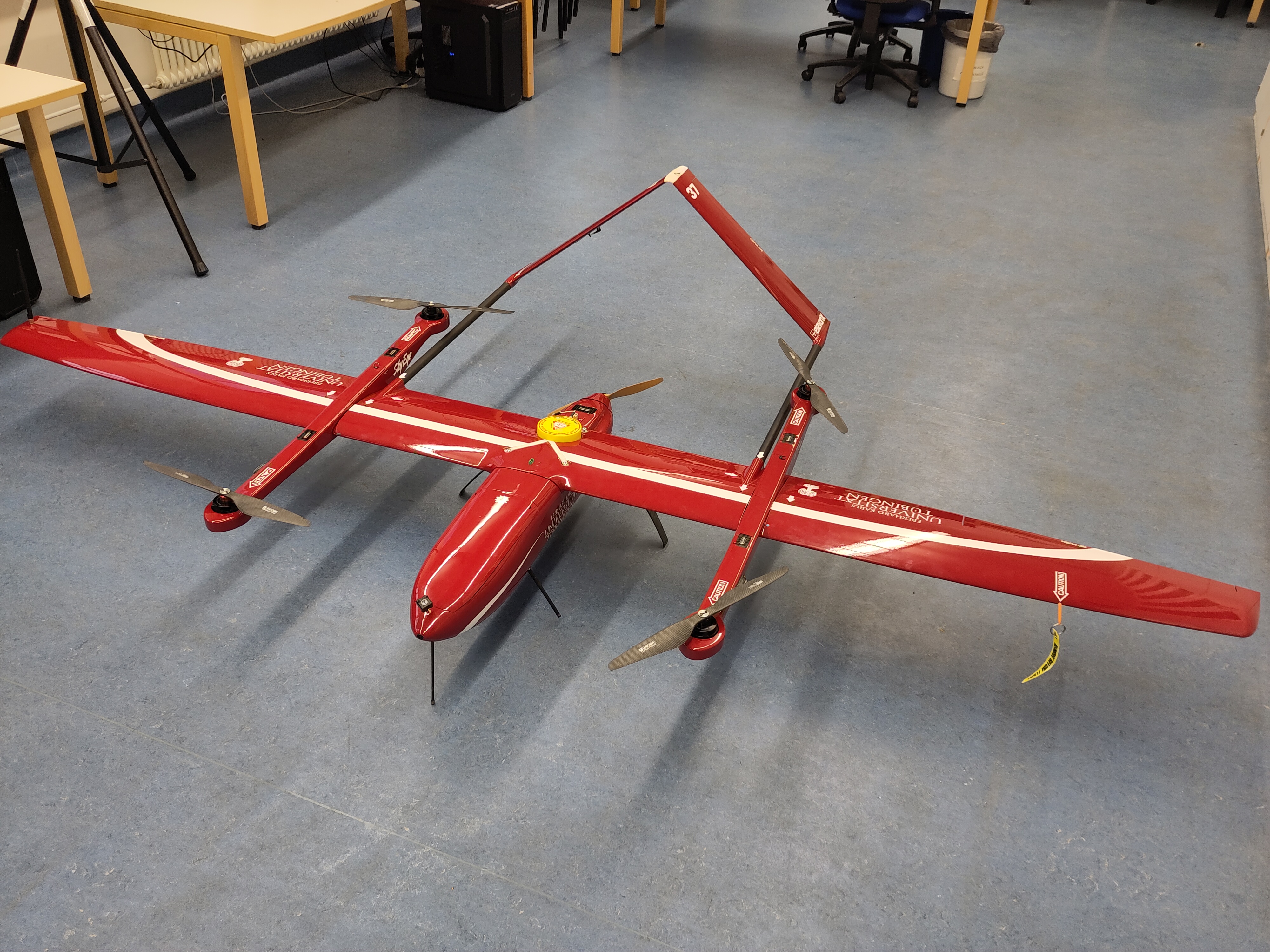}
    \caption{ElevonX SkyEye Sierra VTOL}
    \label{fig:elevonx}
\end{figure}

The ElevonX Sierra SkyEye VTOL (Fig. \ref{fig:elevonx}) combines some of the benefits of both, quadcopters and fixed-wing aircraft. It has a significant endurance of up to 3 hours with electric propulsion, which is essential for extended missions. 
Similar to the Trinity F90+, it is a VTOL UAV and therefore comes with the same advantages and disadvantages compared to a quadcopter. The cruise speed and range are well-suited for both detailed search operations and extensive area coverage.

The adjustability of the ElevonX drone is particularly beneficial for maritime search and rescue. While we only used the the electric propulsion in our experiments, the availability of gas propulsion is a nice addition for operational deployment as it offers the longest endurance for any of the tested drones. However, the vibrations a combustion engine produces may hamper image quality.

One drawback of the Sierra VTOL is its reduced user-friendliness, as it demands continual practice from operators due to its complex nature. Additionally, the system requires two operators to function effectively, which ties up limited resources.

The payload versatility and capacity of $3$ kg mean that the drone can be equipped with various sensors, such as optical and thermal cameras, to capture a wide range of data during both day and night operations. In particular, we were able to install an NVIDIA Jetson Orin NX on the drone and connect it to our RGB camera which enabled us to perform onboard processing for Region of Intereset proposals, as discussed later. The Sierra VTOL allows for up to $120$ Watts of power within the drone.

In summary, the DJI M210 is ideal for targeted search operations and detailed inspections, the Trinity F90+ excels in extensive area mapping and data generation, and the ElevonX Sierra SkyEye VTOL offers a blend of endurance, adjustability, and payload capacity while being the most complex to operate. The choice of drone would depend on the specific phase of the mission and the mission itself. 
The critical considerations in this context are: is onboard processing desired? Is there a need to capture data extensively? Is there a focus on continuously monitoring small, specific areas, or is the objective to search larger spaces?

\section{Software Solution for SAR Drones}
\label{sec:software}

Our proposed software solution is divided into two components: the software operating on the embedded GPU onboard the UAV, and the program running on the ground station. In the following, we provide a concise description of the functionalities. The software, containing both parts, will be made publicly available upon acceptance for publication.

\subsection*{Onboard UAV Software}
The software onboard the UAV assumes that the flight controller and path planning system are taken care of. For example, with the UAVs discussed in section \ref{sec:drones} these functionalities are provided by the manufacturer. This software has, in essence, three main tasks:

1) Running the camera: The software fetches the video stream from the connected camera and then preprocesses this video data for both detection and streaming.
2) Region of Interest (RoI) proposal: Detect regions of interest for closer examination by the ground station software. The identified regions will be transmitted to the ground station in full detail. However, this process might be constrained by the available bandwidth.
3) Data Transmission: Establishing a connection with the ground station software is crucial. The onboard software streams the down-scaled full image, the RoIs in full quality, and metadata about these RoIs. This metadata includes the time stamp to align the RoIs with the video stream.
Note: The actual streaming mechanism is a topic for potential further research work but falls outside the scope of this work. Some possible solutions involve streaming FPGAs implementing a dedicated codec for high-quality RoIs embedded in a low-quality video stream \cite{stabernack2021architecture_fpga}.

\subsection*{Ground Station Software}

The receiving part of the software assumes the ground station to be a GPU server equipped with powerful graphics cards to run demanding object detection models. This way, the software can run these detectors on the received RoIs. Essentialy this software needs to perform four tasks:

1) Data Streaming: This software needs to establish and maintain a connection with the UAV software, receiving the various data streams sent by it.

2) Detailed Detection: It runs powerful object detectors on the RoIs proposed and streamed by the drone software. For this, the software in the ground station needs to manage the workload imposed by multiple RoIs and over multiple video frames across all available GPUs.

3) Operator Interface: The software presents the information in a GUI to an operator at the ground station, enabling them to take action if required, such as identifying individuals in distress.

4) Custom RoI Requests: Additionally, the operator has the ability to request custom RoIs if they believe the drone software may have missed something. These custom RoIs are then transmitted back to the drone software and then are also transmitted in full quality to the ground station.

\section{Experiments}
\label{sec:experiments}

As object detection is studied extensively already \cite{carion2020detr, girshick2015fastrcnn, wang2023yolov7} and theory and experiments on the path planning of the drone are out of the scope of this work, we confine ourselves to exploring various proposal methods for regions of interest predicted by the embedded GPU on the drone. This requires algorithms to be both, to ensure that no objects present in the scene are overlooked.

\subsection*{Region of Interest Proposer Methods}

\begin{figure*}[ht]
    \begin{center}
        \includegraphics[width=0.49\linewidth]{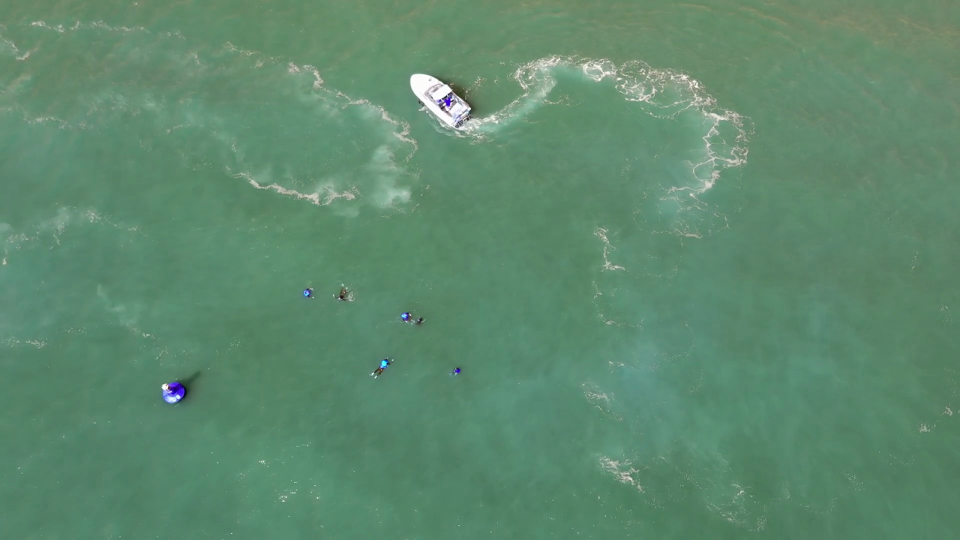}
        \includegraphics[width=0.49\linewidth]{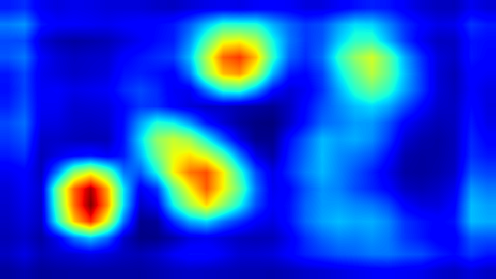}
    \end{center}
    
\vspace{-3mm}

    \begin{center}
        \includegraphics[width=0.49\linewidth]{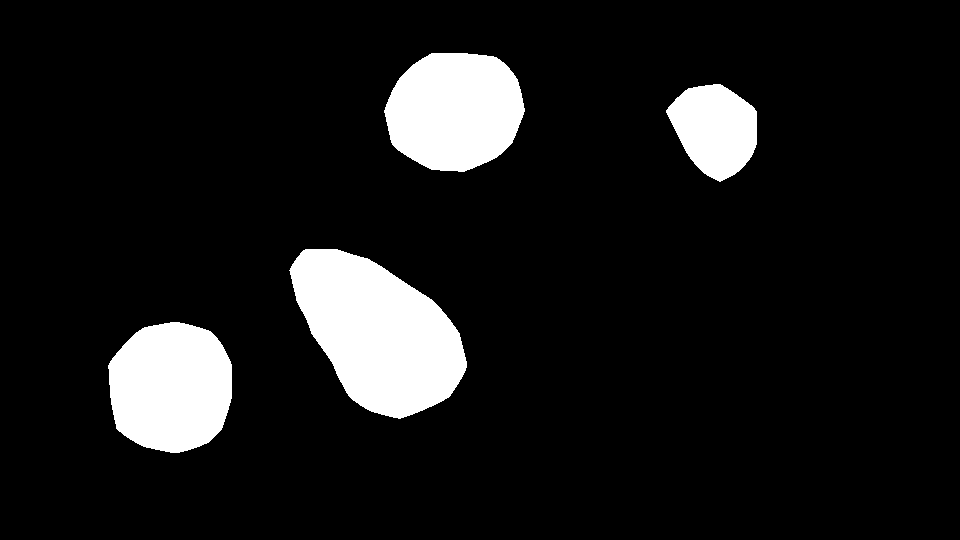}
        \includegraphics[width=0.49\linewidth]{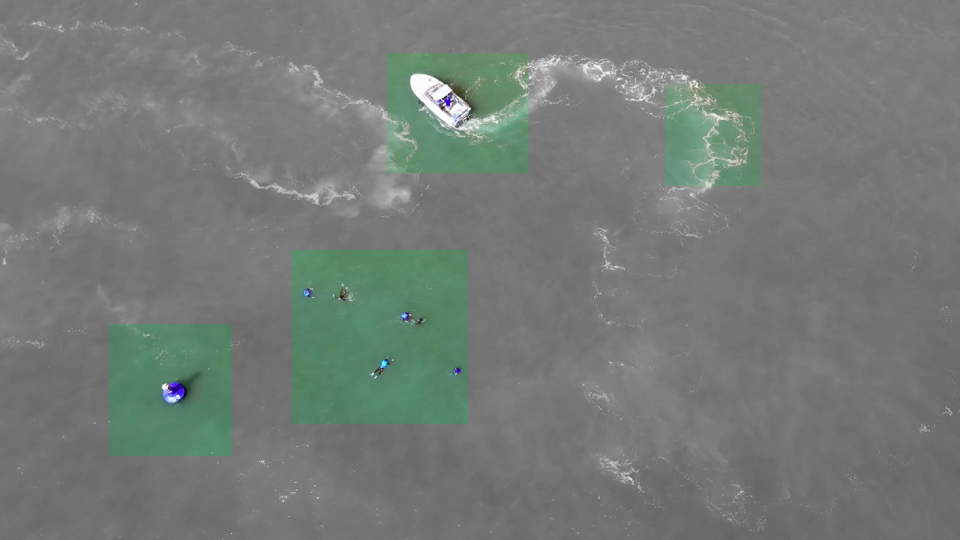}
    \end{center}
    \caption{Example of a heat map produced by our saliency detection method. The image on the top left is from the test set of the SeaDronesSee data set \cite{varga2022seadronessee}. The image on the top right shows the corresponding heat map. The bottom left shows the resulting binary map after rounding each pixel to $0.$ and $1.$. The bottom right shows the resulting RoIs on the images, the background is in gray scale to highlight the predictions. We can see, that all relevant instances in the image are detected. From left to right, the bounding boxes detect a buoy, swimmers, a boat, and waves -- the only irrelevant detection.}
    \label{fig:saliency_example}
\end{figure*}

The first method in this work's evaluation for Region of Interesent (RoI) proposals is hugely based on Grad-CAM++\cite{chattopadhay2018grad}. Its goal it to enhance the understanding of how deep learning models, particularly CNNs, process and interpret visual data. The authors do so by generating visual explanations, so called \textit{saliency maps}, for the predictions made by Convolutional Neural Networks. Figure \ref{fig:saliency_example} shows an example of a saliency map.

In essence, this technique creates a heat map on the image, where pixels with higher values indicate areas that the neural network relies on more for its decision-making process. This is done by calculating higher order derivatives of the network's classification output in relation to specific pixels. Intuitively, this approach makes sense; examining how altering a particular pixel influences the network's class output of a specific bounding box gives insight into the significance of that pixel in the network's decision-making process.

While originally developed for shedding light on the decision-making process of convolutional neural networks, we adapt this method in our study for class-agnostic object detection. We achieve this by processing the output heatmap and rounding each value to either $0$ or $1$, resulting in a binary map. Next, OpenCV's\cite{opencv_library} \texttt{findContours}\cite{suzuki1985topological} function, traces the outlines of the shapes in the image that correspond to the regions marked by the value $1$ in the binary map. Since the \texttt{findContours} function is furthermore able to fill the interior of this outlined shape, it can recover bounding boxes by simply fitting the smallest possible axis-aligned rectangular box around each connected component of the areas marked as $1$ in this binary heat map. For this approach we merely employ a generic purpose backbone frequently used for computer vision tasks, ResNet-18 \cite{he2016deepres}.

The primary benefit of this method lies in its class-agnostic nature. Given that the model isn't specifically trained on maritime data, applying it to out-of-distribution scenarios poses no issue. This flexibility allows for broader applicability across various data, even if it differs significantly from the model's training data. In the following we call this approach \textit{saliency detection}. Figure \ref{fig:saliency_example} shows a graphical example of this algorithm.

The second method we employ for RoI proposals is YOLOv7 \cite{wang2023yolov7}. Since this approach is a vanilla generic object detector, it needs training on data that is representative of the same distribution as the test data. This is an obvious drawback in comparison to the saliency detection method.

However, this method ranks among the most efficient and accurate object detectors currently available to the research community. This model consistently ranks as one of the top-performing models in relevant benchmarks for real-time object detection\cite{coco_leaderboard}, such as the MS COCO dataset \cite{lin2014microsoft_coco}.
Another significant advantage of YOLOv7 is its user-friendliness. Its implementation is easy to use and operational immediately, requiring only a single line of command for retraining or fine tuning. This feature is particularly beneficial for our application, as the model will be handed over to users who are not AI-experts. They still might need to fine tune the model on newly gathered data.

Since the task differs slightly from standard object detection, we experimented with two different configurations of YOLOv7. In the first, we use it as a conventional object detector, trained on SeaDronesSee in the usual manner\footnote{\url{https://github.com/WongKinYiu/yolov7/blob/main/data/hyp.scratch.p5.yaml}}. For the second configuration, we modified the dataset as follows: every bounding box in the dataset is expanded to a minimum size of $500 \times 500$ pixels (while images are $3840 \times 2160$). This adjustment is based on the assumption that the detection of an individual or object might indicate the presence of additional items in its vicinity, potentially leading to their detection as well. This modification is a significant alteration to bounding boxes for many objects considering that a vast number of them are merely around $20 \times 20$ pixels in size.

This strategy, in theory, might lead to a decrease in detection performance in scenarios with limited streaming bandwidth, but proof advantageous when the available bandwidth is sufficiently high.

\subsection*{Data Set}

We conducted evaluations of our algorithms using the test set from the second iteration of the SeaDronesSee\cite{varga2022seadronessee} dataset. This test set contains of $4235$ images, while the training and validation sets are $8125$ and $1431$ images, respectively.
It is composed of images featuring from $1$ to $15$ swimmers per image, as well as small vessels in open water. These images were captured using multiple different drones, including the DJI M210 and Trinity F90+ models discussed in this work. The dataset also was captured by various cameras and diverse weather and lighting conditions, as it was gathered over multiple days.

\subsection*{Evaluation Metrics}

To accommodate the unique requirements of this paper, we employ specialized evaluation metrics. This involves a slightly modified version of the conventional metrics precision, recall, and the F1 score. These metrics are widely used in the field of computer vision and machine learning to assess the accuracy of predictive models. The necessary modification lies in how a 'True Positive' needs to be correctly defined in this context.

Typically, in the process of matching predicted and ground truth bounding boxes, the \textit{Intersection over Union} (IoU) is calculated for each pair of predictions and ground truth boxes $(p_j, g_k) \in P \times GT$, where $P$ is the set of predicted boxes and $GT$ the set of ground truth boxes. A match is established when a pair yields an IoU greater than $0.5$ and also represents the highest IoU among all possible pairings, formally:
\begin{align}
    \operatorname{IoU}(p_j, g_k) & \geq \frac{1}{2} \\
    \operatorname{IoU} (p_j, g_k) & = \max_{m,n} ~ \operatorname{IoU} (p_m, g_n) \label{eq:one_to_one_condition}
\end{align}
Here, the maximum in line \ref{eq:one_to_one_condition} only is taken over all predictions $p_m$ and ground truth boxes $g_n$ that are not yet matched.
This matching process is one-to-one, linking each ground truth box to at most one predicted box. Each prediction matched to a ground truth box is called a \textit{True Positive}.

This approach is well-suited for precise object detection, where each prediction is expected to correspond to exactly one object within the image. However, for Region of Interest (RoI) prediction, where the goal is to enclose all objects present in the given scene, the matching method needs to be adjusted.

Specifically, this implies that the strict one-to-one correspondence between predictions and ground truth boxes are not necessary, nor is the precise size alignment of the predicted bounding boxes with the actual objects.

Therefore, instead of IoU we use \textit{Intersection over Ground Truth} (IoGT), defined as
\begin{displaymath}
    \operatorname{IoGT} (p, g) := \frac{\operatorname{area} (p \cap g)}{\operatorname{area} (g)} \in [0,1] .
\end{displaymath}

In simpler terms, this term focuses only on the area of the actual object, rather than looking at the area covered by both the predicted and actual object together. This way, even if a predicted area is much larger than needed, it can still be considered a match. In the extreme case of a predicted RoI covering the whole image, every ground truth bounding box would be counted as a true positive.
With the traditional Intersection over Union method, a very large prediction would result in a low value, possibly missing the match.
More formally, similarly to above equations, our matching procedure is:
\begin{align}
    \operatorname{IoGT}(p_j, g_k) & \geq \frac{1}{2} \\
    \operatorname{IoGT} (p_j, g_k) & = \max_{m,n} ~ \operatorname{IoGT} (p_m, g_n) \label{eq:one_to_many}
\end{align}
Here, the maximum in equation \ref{eq:one_to_many} is taken over all $p_m$ and $g_n$, regardless of whether they were matched to some other box already or not. Each ground truth box that matches with a predicted box is now considered a true positive.

Using this altered metric, we strive to achieve a shift in focus from accuracy on individual objects to predicted areas covering all present objects in the scene, even if it's not a perfect match for each one. This is the goal of region of interest prediction. We believe, our metric achieves precisely this goal.

\subsection*{Results}

\begin{figure}[ht]
    \centering
    \includegraphics[width=0.99\linewidth]{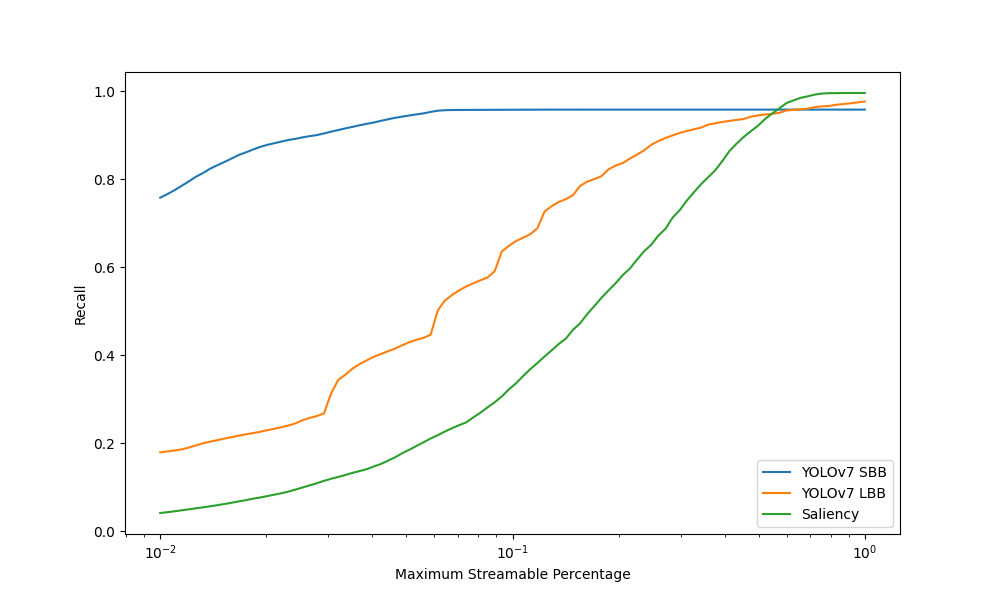}
    \caption{Logarithmic plot of the recall on the SeaDronesSee \cite{varga2022seadronessee} test set for saliency detection and two different versions of YOLOv7 \cite{wang2023yolov7}. The percentage of the maximum streamable resolution in relation to the full resolution is plotted on the $x$-axis.}
    \label{graph:recall}
\end{figure}

\begin{figure}[ht]
    \centering
    \includegraphics[width=0.99\linewidth]{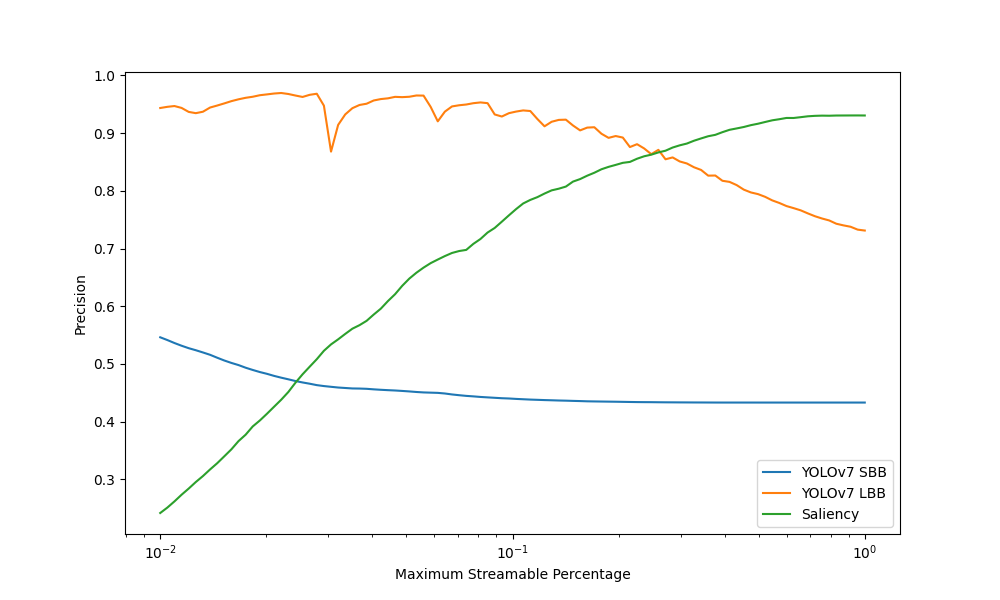}
    \caption{Logarithmic plot of the precision on the SeaDronesSee test set for saliency detection and the discussed versions of YOLOv7. The percentage of the maximum streamable portion of the full resolution is plotted on the $x$-axis.}
    \label{graph:precision}
\end{figure}

\begin{figure}[ht]
    \centering
    \includegraphics[width=0.99\linewidth]{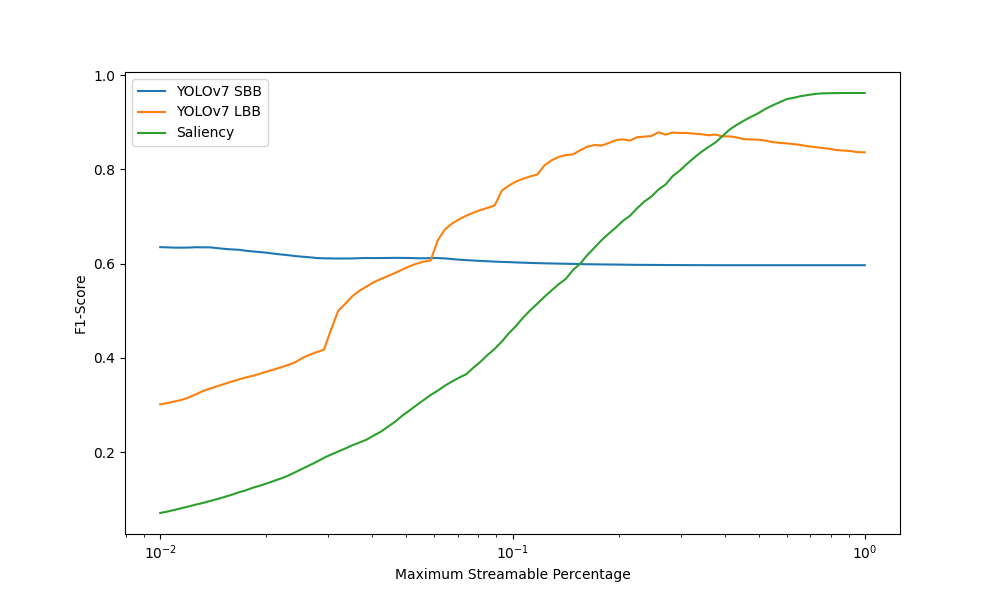}
    \caption{Logarithmic plot of the F1-score on the test set of the SeaDronesSee data set for saliency detection and two different versions of YOLOv7. The percentage of the maximum streamable portion of the full resolution is plotted on the $x$-axis.}
    \label{graph:f1}
\end{figure}

To assess the performance of the discussed region proposal methods, we compute their adapted precision, recall, and F1-score with the discussed modifications. Additionally, we simulate a scenario with reduced bandwidth for the data link from the drone to the ground station. In our evaluation, this is achieved by limiting the predictions generated by the algorithm to the quantity that can feasibly be transmitted given the bandwidth constraints. We do so as follows: given a portion $r \in [0,1]$, which denotes the maximum portion that can be streamed in full quality, and predictions $P = \{ p_1, \ldots , p_n \}$. Then, in the case of saliency detection, where we do not have information about the quality of the boxes, we chose a subset $\{ p_{k_1}, \ldots , p_{k_m} \} \subset P$ such that $\operatorname{area} \big( p_{k_1} \cup \ldots \cup p_{k_m} \big)$ is maximal and not greater than $r$. For YOLOv7, where each bounding box comes with a confidence score, we chose the boxes with the highest score until the area exceeds $r$.
In either case, once the bandwidth limitation is reached, we select the next bounding box in this order. For saliency detection, we opt for the largest remaining box, assuming that it represents the highest quality detection. We fit the largest axis-aligned box (with equal center) into the selected box, while still accounting for the streaming restriction, and append this box to the subset. Figures \ref{graph:recall}, \ref{graph:precision}, and \ref{graph:f1} show the results of the performance under these conditions. There, the maximum streamable percentage of the image is given by the $x$-axis while the result of the algorithm in the respective metric is plotted on the $y$-axis.

This approach allows us to understand how the methods perform under restricted data transmission conditions.

Figure \ref{graph:recall} shows the recall of the compared algorithms, arguably the most important metric for RoI proposal methods in mSAR missions. We observe that vanilla YOLOv7 outperformed the other models in almost all scenarios with bandwidth limitations, making it the preferred choice under heavy bandwidth limitations. However, as the bandwidth restriction vanishes, meaning $r \rightarrow 1$, the YOLO variant trained on the modified data set with larger bounding boxes begins to excel, surpassing the vanilla version.
Similarly but even more so, the saliency detection method surpasses both YOLOv7 variants once the data link allows for around $60\,\%$ or more of the image to be streamed. This is not surprising, since the YOLO methods are trained to predict few false positives.

In addition, saliency detection has a unique benefit: it maintains consistent performance even when applied to data it wasn't trained on, unlike its counterparts, which were explicitly trained on SeaDronesSee.
An interesting observation can be made regarding the YOLO model with enlarged bounding boxes: the noticeable jumps upwards in the graph occur when the bandwidth allows for one additional bounding box, leading to more additional detections quickly. These jumps are also visible in figures \ref{graph:precision} and \ref{graph:f1}. In graph \ref{graph:precision} they look like these additional bounding boxes harmed performance but figure \ref{graph:f1} shows, that they are a net-gain for overall detection performance. 
The superiority of saliency detection for a high bandwidth can also be seen in the precision and F1-score plots, figures \ref{graph:precision} and \ref{graph:f1}. Here, already when roughly $30\,\%$ of the full image can be streamed, the figures show that saliency detection overtakes the other two algorithms in performance. Although a high recall is the most important aspect for RoI proposal methods in mSAR missions, a high precision might as well be beneficial, as it reduces the burden on the object detector or the human operator present at the ground station.

\subsection*{Speed Benchmarks}
\label{sec:speed_benchmarks}

In order to evaluate the possibility of onboard processing for the investigated algorithms, we conducted speed benchmarks on an NVIDIA Jetson Orin \cite{nvidia_orin}. We averaged running times for both algorithms on the test set of the SeaDronesSee benchmark data set.
We used the TensorRT framework \cite{vanholder2016efficient_tensorrt} to optimize the ResNet and the subsequent heat map generation in the saliency detection algorithm as discussed above. This part achieved an average processing speed of $51.0$ frames per second (FPS) on the Orin. The post processing generating bounding boxes, employing OpenCV \cite{opencv_library}, averaged $30.1$ FPS. Running them in sequence hence results in roughly $18.9$ FPS, falling short of running in real time. We conclude that, since the tasks are naturally separate, splitting the two tasks on two different processing units, one dedicated GPU for the neural forward pass and heat map generation and one dedicated CPU for the OpenCV computations, would achieve real-time by being as fast as the slower of the two operations, resulting in $30.1$ FPS.
On the other hand, YOLOv7 achieved an average of $45.5$ FPS on the NVIDIA Jetson Orin while running natively in PyTorch \cite{NEURIPS2019_9015_pytorch}. Although there is potential for further acceleration through TensorRT optimization, we deemed it unnecessary since it already achieved real-time performance. Since both researched YOLOv7 methods only differ in the training of the network, inference speed is equal for both.

\section{Conclusion}
\label{sec:conclusion}

To summarize, our research highlights some of the features and caveats about using small fixed-wing drones and quadcopters in maritime search and rescue (mSAR) operations. Furthermore, we developed and implemented a specialized software framework specifically tailored for mSAR missions. This framework allows for models predicting regions of interest (RoIs) onboard the UAVs and also using more powerful object detectors at a ground stations, thereby reducing the cognitive and visual load on operators. Additionally, we evaluated various RoI proposal methods specifically focusing on different bandwidth constraints, thereby highlighting their practicality in real-world mSAR scenarios involving UAVs.
\newpage
\pagebreak

{\small
\bibliographystyle{ieee_fullname}
\bibliography{egbib}
}

\end{document}